\pdfoutput=1

\documentclass[11pt]{article}

\usepackage[preprint]{acl}

\usepackage{times}
\usepackage{latexsym}

\usepackage[T1]{fontenc}

\usepackage[utf8]{inputenc}

\usepackage{microtype}

\usepackage{inconsolata}

\usepackage{graphicx}
\usepackage{amsmath}

\usepackage{algorithmic,algorithm}
\renewcommand{\algorithmiccomment}[1]{\bgroup\hfill$\triangleright$~#1\egroup}
\usepackage[linguistics]{forest} 
\usepackage{synttree}
\usepackage{subcaption}
\usepackage{tikz-dependency}
\usepackage[cjk]{kotex}

\usepackage{langsci-gb4e}

\microtypesetup{protrusion=false} 

\title{Proposing TAGbank as a Corpus of Tree-Adjoining Grammar Derivations}

\author{
Jungyeul Park\\
Department of Linguistics\\
The University of British Columbia\\
\texttt{jungyeul@mail.ubc.ca} \\
}

\begin{document}
\maketitle

\begin{abstract}
The development of lexicalized grammars, particularly Tree-Adjoining Grammar (TAG), has significantly advanced our understanding of syntax and semantics in natural language processing (NLP). While existing syntactic resources like the Penn Treebank and Universal Dependencies offer extensive annotations for phrase-structure and dependency parsing, there is a lack of large-scale corpora grounded in lexicalized grammar formalisms. To address this gap, we introduce TAGbank, a corpus of TAG derivations automatically extracted from existing syntactic treebanks. 
This paper outlines a methodology for mapping phrase-structure annotations to TAG derivations, leveraging the generative power of TAG to support parsing, grammar induction, and semantic analysis. 
Our approach builds on the work of CCGbank, extending it to incorporate the unique structural properties of TAG, including its transparent derivation trees and its ability to capture long-distance dependencies. We also discuss the challenges involved in the extraction process, including ensuring consistency across treebank schemes and dealing with language-specific syntactic idiosyncrasies. Finally, we propose the future extension of TAGbank to include multilingual corpora, focusing on the Penn Korean and Penn Chinese Treebanks, to explore the cross-linguistic application of TAG’s formalism. By providing a robust, derivation-based resource, TAGbank aims to support a wide range of computational tasks and contribute to the theoretical understanding of TAG’s generative capacity.
\end{abstract}

\section{Introduction}

The development of syntactic resources has played a pivotal role in advancing natural language processing (NLP), particularly in parsing and grammar-based applications. While widely used resources such as the Penn Treebank \cite{marcus-santorini-marcinkiewicz-1993-building,taylor-marcus-santorini:2003} and Universal Dependencies \cite{de-marneffe-etal-2021-universal} provide valuable annotations for phrase-structure and dependency syntax, there remains a lack of large-scale corpora grounded in lexicalized grammar formalisms. To address this gap, we introduce TAGbank, a corpus of TAG \citep{joshi-levy-takahashi-1975-tree,joshi-1985-how,joshi-1987-introduction} derivations derived from existing syntactic annotations.

Lexicalized grammars such as TAG and Combinatory Categorial Grammar (CCG) \citep{steedman:2000,steedman-baldridge:2011} offer a richer interface between syntax and semantics compared to more traditional context-free grammar (CFG)-based approaches. Unlike Probabilistic CFGs (PCFGs), which rely heavily on constituent structure with limited lexical information, lexicalized formalisms encode dependencies through lexically anchored elementary structures. TAG, in particular, offers extended domain of locality \citep{joshi-levy-takahashi-1975-tree} and a transparent derivation structure \citep{joshi-schabes-1997-tree}, making it suitable for capturing long-distance dependencies \citep{joshi-1985-how} and compositional semantics \citep{joshi-vijayshanker-2001-compositional}.

TAGbank is building on the precedent set by CCGbank \cite{hockenmaier-steedman-2007-ccgbank}, which translates the Penn Treebank into a corpus of CCG derivations. Building on this idea, our work extracts TAG derived and derivation structures from phrase-structure and dependency treebanks, resulting in a new syntactic resource that supports robust, cross-linguistic grammar-based processing and semantic interpretation. TAGbank is designed to bridge the gap between existing syntactic resources and lexicalized grammar formalisms, while also facilitating grammar induction, parser evaluation, and formal grammar research.

The contributions of this paper are twofold. First, we propose a methodology for deriving TAG structures from existing syntactic annotations, enabling the systematic transformation of phrase-structure representations into TAG derivations. Second, in the course of this transformation, we identify and analyze key linguistic and computational challenges inherent in the extraction process, including issues related to lexical anchoring, structural decomposition, and formal consistency with TAG’s derivational requirements.

\section{Background and Related Work}

TAG was first introduced by \citet{joshi-levy-takahashi-1975-tree} and later formalized as a grammar formalism that balances computational tractability with linguistic expressiveness \citep{joshi-1985-how}. A TAG is defined as a five-tuple ($\Sigma$, $NT$, $I$, $A$, $S$), where $\Sigma$ is a finite set of terminal symbols, $NT$ is a set of nonterminal symbols, $I$ is a finite set of initial trees, $A$ is a finite set of auxiliary trees, and $S$ is a distinguished start symbol.

The sets $I$ and $A$ together constitute the elementary trees of the grammar—the core building blocks from which larger syntactic structures are derived. Initial trees ($\alpha$ trees) represent the predicate-argument structure of lexical items, typically encoding subcategorization frames of verbs or other syntactic heads. They are complete, well-formed trees except for designated frontier nodes marked for substitution. Auxiliary trees ($\beta$ trees), by contrast, capture recursive or modifying constructions such as adverbial adjuncts, relative clauses, or topicalization. Each auxiliary tree contains a distinguished foot node, which is identical in label to its root, thereby allowing adjunction into other trees.\footnote{Because each elementary tree is anchored to a lexical item and encodes its argument structure locally, TAG naturally represents long-distance dependencies and complex syntactic constructions. This property has made TAG particularly effective for analyzing phenomena such as wh-movement, relativization, and coordination, while maintaining a clear connection between syntax and semantics.}

TAG operates with two core tree-composition mechanisms: substitution, which inserts an initial tree at a frontier node labeled by a nonterminal, and adjunction, which inserts an auxiliary tree at an internal node where the root and foot labels match. These operations support the construction of both the derived tree (representing the surface syntactic structure) and the derivation tree (which records the hierarchical history of tree combinations). The derivation tree is particularly important in TAG, as it closely corresponds to both syntactic dependencies and compositional semantic interpretation.

A central linguistic motivation for TAG lies in its ability to capture long-distance dependencies and to model predicate-argument relations within its elementary trees, which is made possible by its extended domain of locality \citep{kroch-joshi-1985-linguistic, kroch-1987-unbounded}. This property allows subcategorization frames and argument structure to be encoded directly within each elementary tree, yielding a highly lexicalized view of grammar. Importantly, the derivation tree in TAG encodes a transparent syntactic history that corresponds closely to compositional semantic interpretation \citep{joshi-schabes-1997-tree}. Each node in the derivation tree is annotated with a tree address that specifies the location where substitution or adjunction occurred—excluding the root—enabling a tight coupling between syntactic composition and semantic interpretation.

From a formal perspective, TAG has been extensively studied for its mathematical properties. TAG is a mildly context-sensitive grammar (MCSG) formalism that admits polynomial-time parsing complexity \citep{vijayshankar-joshi:1986:HLT}. Its generative capacity exceeds that of context-free grammars (CFGs) while remaining computationally tractable, making it well-suited for modeling a range of natural language phenomena.
Subsequent work has clarified TAG’s position within the broader class of mildly context-sensitive formalisms. It has been shown to be closely related in generative power to Linear Indexed Grammars and Head Grammars, while also distinct from them in structural assumptions and derivational mechanisms \citep{joshi-vijayshanker-weir-1991-convergence, vijayshanker-weir-1994-equivalence}. CCG also falls within the mildly context-sensitive class and provides an alternative lexicalized approach to capturing unbounded dependencies and argument structure. While TAG and CCG differ in their formal mechanisms—tree adjunction versus combinatory function application—they are often compared in terms of their expressive power and suitability for computational parsing and grammar induction.
Together, these studies have solidified TAG’s theoretical foundations and emphasized its strength in capturing cross-serial dependencies, long-distance movement, and other constructions that challenge the descriptive adequacy of CFG-based models.

On the practical side, several efforts have aimed to develop TAG grammars from syntactically annotated corpora. The XTAG Project \citep{joshi-2001-xtag} introduced a wide-coverage English grammar based on Feature-based TAG (FTAG), incorporating detailed morphosyntactic constraints through unification-based features. Subsequent work \citep{xia-2001-thesis, chen-vijayshanker:2000, chen:2001, xia-palmer-vijayshanker:2005} explored automatic grammar extraction from treebanks, demonstrating how TAG-style elementary trees can be induced from phrase-structure corpora such as the Penn Treebank. These efforts laid the foundation for semi-automatic construction of TAG grammars and provided empirical frameworks for evaluating them against naturally occurring syntactic data.

Additionally, CCGbank \citep{hockenmaier-steedman-2007-ccgbank}, a transformation of the Penn Treebank into Combinatory Categorial Grammar derivations, exemplifies how phrase-structure data can be reinterpreted in lexicalized, derivational frameworks. Its success as a benchmark for parser evaluation and grammar induction motivates similar efforts for TAG.

While treebanks such as the Penn English Treebank \citep{marcus-santorini-marcinkiewicz-1993-building},  the Penn Chinese Treebank \citep{xue-EtAl:2005}, and the Penn Korean Treebank \citep{han-EtAl:2002} have facilitated research on constituency and dependency parsing across languages, there remains a notable lack of corpus resources that capture the derivational structure of lexicalized grammar formalisms such as TAG.

Our work seeks to fill this gap by proposing the construction of TAGbank—a corpus of TAG derivations automatically extracted from existing syntactic treebanks. In doing so, we adapt the methodology of CCGbank to the structural and formal requirements of TAG, while also extending the resource to typologically diverse languages such as Chinese and Korean. TAGbank offers both a practical dataset for lexicalized grammar parsing and a platform for exploring the linguistic, formal, and cross-linguistic dimensions of TAG.

\section{TAG Derivation Structures}

In TAG, derivations produce two complementary structures: the derived tree, which reflects surface constituency, and the derivation tree, which records the sequence of elementary tree combinations. The derivation tree captures both syntactic dependencies and compositional semantics in a transparent and structured way.

\begin{figure*}
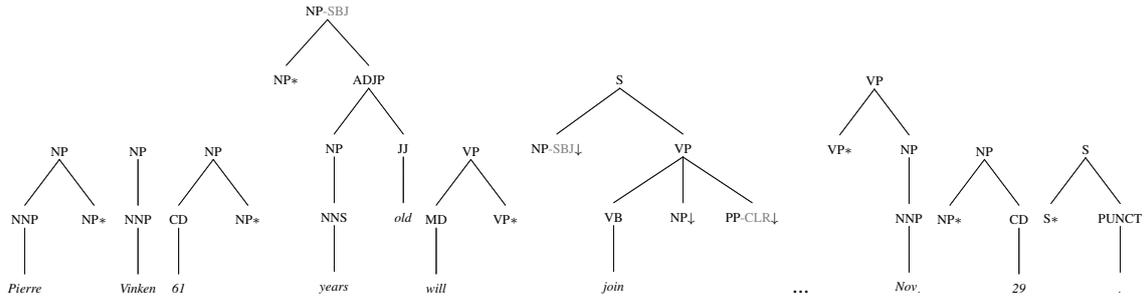

\centering
\resizebox{.95\textwidth}{!}
{
\scriptsize{
{\synttree
[NP [NNP [\textit{Pierre}]] [NP$*$]] 
}~~
{\synttree
[NP  [NNP [\textit{Vinken}]] ]
}~~
{\synttree
[NP [CD [\textit{61}]] [NP$*$]] 
}~~
{\synttree
[NP{\color{gray}-SBJ} [NP$*$]  [ADJP [NP [NNS [\textit{years}] ]] [JJ [\textit{old}]]] ]
}~~
{\synttree
[VP   [MD [\textit{will}]] [VP$*$]]
}~~
{\synttree
[S [NP{\color{gray}-SBJ}$\downarrow$] [VP [VB [\textit{join}]] [NP$\downarrow$] [PP{\color{gray}-CLR}$\downarrow$] ]] 
}~~
{\large ... }~~
{\synttree
[VP [VP$*$] [NP [NNP [\textit{Nov$_{.}$}]]]]
}~~
{\synttree
[NP  [NP$*$] [CD [\textit{29}]] ]
}~~
{\synttree
[S  [S$*$] [PUNCT [$_{.}$]] ]
}
}
}
    \caption{Elementary trees in TAG}
    \label{elementary-trees}
\end{figure*}

Figure~\ref{elementary-trees} presents a sequence of elementary trees extracted for TAG derivations from a sentence corresponding to the Penn Treebank structure in Figure~\ref{penn-treebank}. These trees were automatically induced from phrase-structure annotations using our treebank-to-TAG conversion pipeline (Section~\ref{conversion-process}). Each elementary tree represents a lexically anchored syntactic unit—either an initial tree encoding core argument structure or an auxiliary tree capturing modifiers, adjuncts, or recursive constructions. The figure shows partial trees from the sentence, with key TAG-specific elements indicated using specialized notation.

\begin{itemize}  \setlength\itemsep{0em}
\item The asterisk notation (e.g., \textit{NP$*$}, \textit{VP$*$}, \textit{S$*$}) indicates the foot node of an auxiliary tree, where adjunction can occur.
\item The downward arrow (e.g., \textit{NP$\downarrow$}, \textit{PP{\color{gray}-CLR}$\downarrow$}) marks a substitution site, representing positions where another initial tree can be inserted.
\item Grayed function tags (e.g., \textit{NP{\color{gray}-SBJ}}, \textit{PP{\color{gray}-CLR}}) preserve Penn Treebank grammatical function labels and can be used to track argument roles and semantic structure.\footnote{Previous work on extracting TAG grammars from the Penn Treebank did not preserve function labels in elementary trees, focusing instead on structural decomposition \citep[e.g.,][]{xia-2001-thesis, chen:2001}.}
\end{itemize}

These examples illustrate how surface-level phrase structures (such as \textit{Pierre}, \textit{Vinken}, \textit{61 years old}, \textit{will join}, and \textit{Nov. 29}) are decomposed into reusable TAG elementary trees. For instance, the tree anchored by \textit{join} is an initial tree with substitution sites for its subject (\textit{NP$\downarrow$}), object (\textit{NP$\downarrow$}), and adjunct (\textit{PP{\color{gray}-CLR}$\downarrow$}), modeling its predicate-argument structure explicitly.

Together, these elementary trees form the building blocks for reconstructing the full derivation tree of a sentence. The derivation process combines these trees via substitution and adjunction to yield the final syntactic structure, while preserving both the hierarchical and lexicalized nature of TAG.

\subsection{Automatic conversion process} \label{conversion-process}

The construction of TAGbank follows a treebank-to-TAG conversion pipeline based on the methodology originally proposed by \citet[Chapter~5]{xia-2001-thesis}. This conversion procedure consists of two primary steps: (1) the extraction of elementary trees from annotated phrase-structure trees, and (2) the composition of these extracted trees into Tree-Adjoining Grammar (TAG) derivations, effectively reconstructing the complete syntactic structure of sentences.

In the initial extraction step, phrase-structure trees (denoted as \texttt{ttrees}) from existing treebanks are systematically decomposed into lexically anchored elementary trees (\texttt{etrees}), categorized as either initial trees or auxiliary trees. This decomposition process involves identifying syntactic heads, arguments, and modifiers within each phrase. A head percolation table \citep{collins:1999} is employed to accurately determine the syntactic head of each phrase, while argument and modifier tables provide structural and functional information necessary to distinguish between arguments and adjuncts. Furthermore, a tagset mapping table ensures consistent lexical anchoring and converts the original syntactic annotations into structurally compatible forms within the TAG framework.

In the subsequent composition step, the extracted elementary trees (\texttt{etrees}) are combined through substitution and adjunction operations, forming derived trees that represent complete syntactic analyses. The associated derivation trees explicitly document the hierarchical operations involved, preserving a clear record of each substitution and adjunction event. Notably, these derivation trees align naturally with dependency annotations provided in the Penn Treebank, establishing an explicit and interpretable connection between traditional phrase-structure syntax and TAG’s derivational semantics.\footnote{Attachment ambiguity occurs when an auxiliary tree can be adjoined at multiple potential nodes in a sentence, even when the elementary trees for each word are fixed. Ordering ambiguity arises from multiple possible derivational histories, which may result in distinct derivation trees or even identical derived trees with differing semantic interpretations. An oracle, however, can precisely determine where to adjoin auxiliary trees (resolving attachment ambiguity) and decide the optimal order of adjunction and substitution operations (resolving ordering ambiguity). Thus, we derive an oracle from the original tree to directly resolve both attachment and ordering ambiguities, ensuring optimal decisions at each composition step.}

This automated pipeline facilitates the efficient and systematic transformation of existing syntactic treebanks into a comprehensive corpus of TAG derivations, laying a robust foundation for the development of TAGbank as a linguistic resource.

\subsection{Tree structure examples} \label{examples}

\begin{figure*}
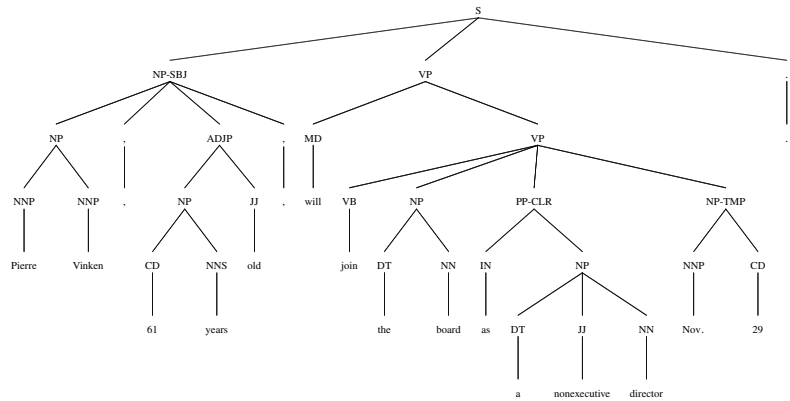
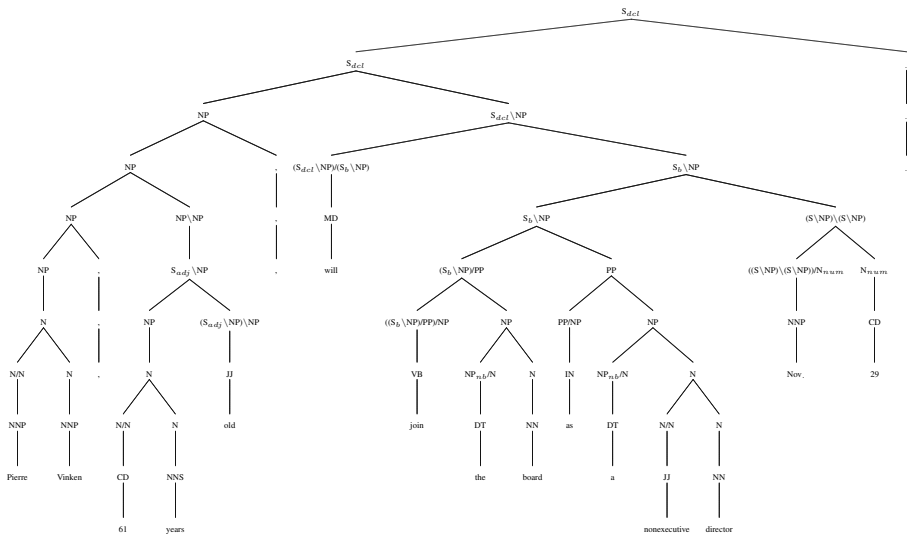
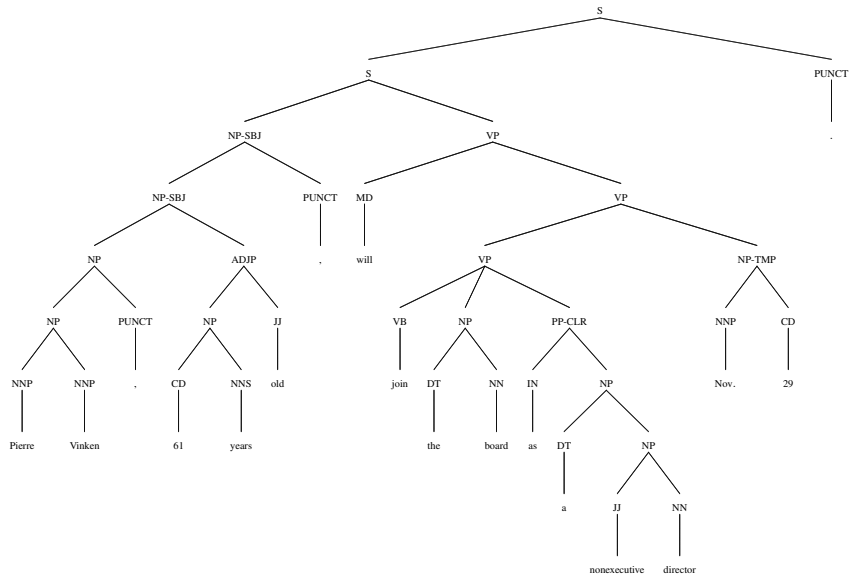

\centering
\begin{subfigure}[b]{\textwidth}    
\centering
\resizebox{.65\textwidth}{!}{
\tiny{
\synttree
[S 
    [NP-SBJ 
      [NP [NNP [Pierre]] [NNP [Vinken]] ]
      [, [,]] 
      [ADJP 
        [NP [CD [61]] [NNS [years]] ]
        [JJ [old]] ]
      [, [,]] ]
    [VP [MD [will]] 
      [VP [VB [join]] 
        [NP [DT [the]] [NN [board]] ]
        [PP-CLR [IN [as]] 
          [NP [DT [a]] [JJ [nonexecutive]] [NN [director]] ]]
        [NP-TMP [NNP [Nov${.}$]] [CD [29]] ]]]
    [${.}$ [${.}$]] ]
}
}
\caption{Penn Treebank representation} \label{penn-treebank}
\end{subfigure}

\hfill

\begin{subfigure}[b]{\textwidth}    
\centering
\resizebox{.75\textwidth}{!}{
\tiny{
\synttree
[ S$_{dcl}$ [ S$_{dcl}$ [ NP [ NP [ NP [ NP [ N [ N/N[NNP[Pierre]] ] [ N[NNP[Vinken]] ] ] ] [ ,[,[,]] ] ] [ NP$\backslash$NP [ S$_{adj}$$\backslash$NP [ NP [ N [ N/N[CD[61]] ] [ N[NNS[years]] ] ] ] [ (S$_{adj}$$\backslash$NP)$\backslash$NP[JJ[old]] ] ] ] ] [ ,[,[,]] ] ] [ S$_{dcl}$$\backslash$NP [ (S$_{dcl}$$\backslash$NP)/(S$_{b}$$\backslash$NP)[MD[will]] ] [ S$_{b}$$\backslash$NP [ S$_{b}$$\backslash$NP [ (S$_{b}$$\backslash$NP)/PP [ ((S$_{b}$$\backslash$NP)/PP)/NP[VB[join]] ] [ NP [ NP$_{nb}$/N[DT[the]] ] [ N[NN[board]] ] ] ] [ PP [ PP/NP[IN[as]] ] [ NP [ NP$_{nb}$/N[DT[a]] ] [ N [ N/N[JJ[nonexecutive]] ] [ N[NN[director]] ] ] ] ] ] [ (S$\backslash$NP)$\backslash$(S$\backslash$NP) [ ((S$\backslash$NP)$\backslash$(S$\backslash$NP))/N$_{num}$[NNP[Nov$_{.}$]] ] [ N$_{num}$[CD[29]] ] ] ] ] ] [ $_{.}$[$_{.}$[$_{.}$]] ] ] 
}
}
\caption{CCGbank representation} \label{ccgbank-tree}
\end{subfigure}

\hfill

\begin{subfigure}[b]{\textwidth}    
\centering
\resizebox{.7\textwidth}{!}{
\tiny{
\synttree
[S [S
    [NP-SBJ [NP-SBJ 
      [NP [NP [NNP [Pierre]] [NNP [Vinken]] ] [PUNCT [,]] ]
      [ADJP 
        [NP [CD [61]] [NNS [years]] ]
        [JJ [old]] ] ] [PUNCT [,]] 
       ]
    [VP [MD [will]] 
      [VP [VP [VB [join]] 
        [NP [DT [the]] [NN [board]] ]
        [PP-CLR [IN [as]] 
          [NP [DT [a]] [NP [JJ [nonexecutive]] [NN [director]] ]]] ]
        [NP-TMP [NNP [Nov${.}$]] [CD [29]] ]]] ]
    [PUNCT [${.}$]] ]
}
}
\caption{TAGbank representation (proposed)} \label{tagbank-tree}
\end{subfigure}

\caption{Syntactic representations of the same sentence in Penn Treebank, CCGbank, and TAGbank} \label{example-of-trees}
\end{figure*}

Figure~\ref{example-of-trees} illustrates how the same sentence in \eqref{pierre} is represented across three different syntactic formalisms: phrase-structure (Penn Treebank), Combinatory Categorial Grammar derivations (CCGbank), and Tree-Adjoining Grammar derivations (TAGbank). These representations reflect differing assumptions about syntactic composition, lexicalization, and derivational transparency, which are central to our motivation for constructing TAGbank.

\begin{exe} 
\ex \label{pierre} \textit{Pierre Vinken, 61 years old, will join the board as a nonexecutive director Nov. 29.} 
\end{exe}

\paragraph{Phrase-Structure Representation (Penn Treebank)}
Figure~\ref{penn-treebank} shows the standard phrase-structure tree from the Penn Treebank. This representation captures hierarchical constituent relations using labeled brackets and syntactic categories such as \textit{NP}, \textit{VP}, and \textit{ADJP}. It organizes surface syntax into nested tree structures but lacks explicit encoding of lexical heads or derivational history. While effective for constituency parsing, this format provides limited support for tracing predicate-argument structure or long-distance dependencies, and requires external rules or heuristics to recover derivational semantics or dependency relations.

\paragraph{Lexicalized Derivation in CCG (CCGbank)}
Figure~\ref{ccgbank-tree} presents the same sentence in CCGbank, which translates the Penn Treebank into Combinatory Categorial Grammar derivations. CCGbank encodes both syntactic and semantic composition using rich type-driven combinatory rules and lexicalized categories. Each word is associated with a syntactic category such as \textit{$($S\textsubscript{dcl}\textbackslash NP$)$/NP}, which determines its combinatory behavior. The derivation tree reflects the step-by-step functional application and composition, capturing predicate-argument structure and syntactic dependencies through a fully lexicalized grammar. However, the abstractness and notational density of CCG derivations may hinder readability and cross-linguistic portability.

\paragraph{Lexicalized Derivation in TAG (TAGbank)}
Figure~\ref{tagbank-tree} illustrates the proposed TAGbank representation, which reconstructs TAG derivation trees from phrase-structure annotations. Like CCG, TAG employs lexicalized elementary trees as the core units of syntactic composition. Each lexical item anchors an elementary tree, classified as either an initial tree, which encodes basic argument structure, or an auxiliary tree, which captures modifiers or recursive constructions. The TAG derivation encodes the sequence of substitutions and adjunctions that yield the surface structure. Notably, the derivation tree provides a transparent mapping from syntactic operations to compositional semantics, facilitating interpretation and downstream semantic analysis. In this figure, adjunction is used to insert the auxiliary tree anchored by \textit{old}, while substitution integrates arguments such as \textit{the board} and \textit{a nonexecutive director}.

\paragraph{Comparative Perspective}
These examples highlight key structural distinctions among the three formalisms:

\begin{description} \setlength\itemsep{0em}
\item [Lexicalization] Both CCGbank and TAGbank associate lexical anchors with syntactic structure, while the Penn Treebank does not explicitly encode lexical heads.

\item [Derivational history] CCG and TAGbank retain derivation trees that record the compositional history of syntactic operations, unlike the Penn Treebank, which represents only surface structure.

\item [Surface vs. deep structure] TAGbank preserves constituent labels from the Penn Treebank but augments them with derivational information, offering a balance between human-readable structure and formal derivational detail.

\item [Transparency and semantic alignment] TAG derivation trees align closely with compositional semantics, supporting tasks such as semantic role labeling, scope interpretation, and natural language generation.

\end{description}

The TAGbank representation is designed to combine the structural clarity of Penn Treebank with the derivational expressiveness of lexicalized formalisms. By enabling alignment between surface syntax and derivation structure, TAGbank facilitates both linguistic analysis and computational modeling. In multilingual contexts, this approach provides a unified framework for comparing syntactic derivations across languages with diverse word order, morphology, and syntactic behavior.

\subsection{Proposed TAGbank format}

We propose a novel TAGbank format designed to integrate TAG derivational information into a tabular, token-aligned representation similar to CoNLL-U. This format supports both PTB-style linearizations \citep{vinyals-etal-2015-grammar} and the incorporation of TAG tree family information from the XTAG English Grammar.\footnote{\url{https://www.cis.upenn.edu/~xtag/gramrelease.html}}

The TAGbank format is structured with the following fields: \textsc{idx}, \textsc{lex}, \textsc{pos}, \textsc{hd}, \textsc{elem}, \textsc{rhs}, and \textsc{lhs}. Here:  
\begin{itemize} \setlength\itemsep{0em}
    \item \textsc{idx} denotes the token index.
    \item \textsc{lex} is the surface lexical item.
    \item \textsc{pos} gives the Penn Treebank POS tag.
    \item \textsc{hd} encodes the syntactic head for the token following dependency conventions.
    \item \textsc{elem} specifies the TAG elementary tree type---either an initial tree ($\alpha$) or auxiliary tree ($\beta$). In future releases, this field may contain actual tree names from the XTAG grammar to enable fine-grained grammar-specific annotations.
    \item \textsc{rhs} and \textsc{lhs} encode bracketed PTB-style constituent information, capturing hierarchical structure in linearized form.
\end{itemize}

To support idiomatic expressions and multiword expressions (MWEs), an alternative version of the TAGbank format introduces special handling of lexicalized MWEs using composite indices (e.g., 5-6) and allows elementary trees to be annotated with lexicalized tree names (e.g., $\beta_{\textit{years old}}$). This mirrors the practice in Universal Dependencies \citep{de-marneffe-etal-2021-universal}, enabling flexible treatment of compositional vs. idiomatic constructions.

Figure~\ref{tagbank-format} illustrates the canonical format for TAGbank using standard token-level alignment and $\alpha$/$\beta$ tree types, without explicit MWE handling. 
Figure~\ref{alt-tagbank-format}, by contrast, presents an extended version of the same sentence with explicit multiword expression annotation, where the phrase ``years old'' is treated as a single unit. This version allows future integration of lexicalized elementary trees from XTAG and potentially enables parallel derivation structures aligned with semantic role annotations.
Both formats are designed to be machine-readable, human-readable, and extensible to future syntactic-semantic annotation layers for TAG parsing and grammar engineering tasks.

\begin{figure*}
\centering
\begin{subfigure}[b]{\textwidth}    
\centering
\footnotesize{
\begin{tabular}{r l l r l l l}
\textsc{idx} & \textsc{lex} & \textsc{pos} & \textsc{hd} & \textsc{elem} & \textsc{rhs} & \textsc{lhs} \\
\hline
1  & \texttt{pierre}       & \texttt{nnp}   & 2  & \texttt{beta}  & \texttt{(S (S (NP-SBJ (NP-SBJ (NP (NP}    & \_ \\
2  & \texttt{viken}        & \texttt{nnp}   & 9  & \texttt{alpha} & \_                                           & \texttt{)NP} \\
3  & \texttt{,}            & \texttt{punct} & 2  & \texttt{beta}  & \_                                           & \texttt{)NP} \\
4  & \texttt{61}           & \texttt{cd}    & 5  & \texttt{beta}  & \texttt{(ADJP (NP}                         & \_ \\
5  & \texttt{years}        & \texttt{nns}   & 2  & \texttt{beta}  & \_                                           & \texttt{)NP} \\
6  & \texttt{old}          & \texttt{jj}    & 5  & \texttt{beta}  & \_                                           & \texttt{)ADJP )NP-SBJ} \\
7  & \texttt{,}            & \texttt{punct} & 6  & \texttt{beta}  & \_                                           & \texttt{)NP-SBJ} \\
8  & \texttt{will}         & \texttt{md}    & 9  & \texttt{beta}  & \texttt{(VP (MD}                           & \_ \\
9  & \texttt{join}         & \texttt{vb}    & 0  & \texttt{alpha} & \texttt{(VP (VP}                           & \_ \\
10 & \texttt{the}          & \texttt{dt}    & 11 & \texttt{alpha} & \texttt{(NP (DT}                           & \_ \\
11 & \texttt{board}        & \texttt{nn}    & 9  & \texttt{alpha} & \_                                           & \texttt{)NP} \\
12 & \texttt{as}           & \texttt{in}    & 9  & \texttt{alpha} & \texttt{(PP-CLR}                           & \_ \\
13 & \texttt{a}            & \texttt{dt}    & 15 & \texttt{alpha} & \texttt{(NP}                               & \_ \\
14 & \texttt{nonexecutive} & \texttt{jj}    & 15 & \texttt{beta}  & \texttt{(NP}                               & \_ \\
15 & \texttt{director}     & \texttt{nn}    & 12 & \texttt{alpha} & \_                                           & \texttt{)NP )NP )PP-CLR )VP} \\
16 & \texttt{nov.}         & \texttt{nnp}   & 9  & \texttt{beta}  & \texttt{(NP-TMP}                           & \_ \\
17 & \texttt{29}           & \texttt{cd}    & 16 & \texttt{beta}  & \_                                           & \texttt{)NP-TMP )VP )VP )S} \\
18 & \texttt{.}            & \texttt{punct} & 9  & \texttt{beta}  & \_                                           & \texttt{)S} \\
\end{tabular}
}
\caption{Canonical TAGbank format with token-level alignment and elementary tree type annotations} \label{tagbank-format}
\end{subfigure}

\hfill \vfill \hfill

\begin{subfigure}[b]{\textwidth}    
\centering
\footnotesize{
\begin{tabular}{r l l r l l l}
\textsc{idx} & \textsc{lex} & \textsc{pos} & \textsc{hd} & \textsc{elem} & \textsc{rhs} & \textsc{lhs} \\
\hline
1   & \texttt{pierre}        & \texttt{nnp}   & 2   & \texttt{beta}  & \texttt{(S (S (NP-SBJ (NP-SBJ (NP (NP} & \_ \\
2   & \texttt{viken}         & \texttt{nnp}   & 9   & \texttt{alpha} & \_                                         & \texttt{)NP} \\
3   & \texttt{,}             & \texttt{punct} & 2   & \texttt{beta}  & \_                                         & \texttt{)NP} \\
4   & \texttt{61}            & \texttt{cd}    & 5   & \texttt{beta}  & \texttt{(ADJP (NP}                       & \_ \\
5-6 & \texttt{years old}     & \_             & \_  & \_             & \_                                         & \_ \\
5   & \texttt{years}         & \texttt{nns}   & 2   & \texttt{beta}  & \_                                         & \texttt{)NP} \\
6   & \texttt{old}           & \texttt{jj}    & 5   & \_             & \_                                         & \texttt{)ADJP )NP-SBJ} \\
7   & \texttt{,}             & \texttt{punct} & 6   & \texttt{beta}  & \_                                         & \texttt{)NP-SBJ} \\
8   & \texttt{will}          & \texttt{md}    & 9   & \texttt{beta}  & \texttt{(VP (MD}                         & \_ \\
9   & \texttt{join}          & \texttt{vb}    & 0   & \texttt{alpha} & \texttt{(VP (VP}                         & \_ \\
10  & \texttt{the}           & \texttt{dt}    & 11  & \texttt{alpha} & \texttt{(NP (DT}                         & \_ \\
11  & \texttt{board}         & \texttt{nn}    & 9   & \texttt{alpha} & \_                                         & \texttt{)NP} \\
12  & \texttt{as}            & \texttt{in}    & 9   & \texttt{alpha} & \texttt{(PP-CLR}                         & \_ \\
13  & \texttt{a}             & \texttt{dt}    & 15  & \texttt{alpha} & \texttt{(NP}                             & \_ \\
14  & \texttt{nonexecutive}  & \texttt{jj}    & 15  & \texttt{beta}  & \texttt{(NP}                             & \_ \\
15  & \texttt{director}      & \texttt{nn}    & 12  & \texttt{alpha} & \_                                         & \texttt{)NP )NP )PP-CLR )VP} \\
16  & \texttt{nov.}          & \texttt{nnp}   & 9   & \texttt{beta}  & \texttt{(NP-TMP}                         & \_ \\
17  & \texttt{29}            & \texttt{cd}    & 16  & \texttt{beta}  & \_                                         & \texttt{)NP-TMP )VP )VP )S} \\
18  & \texttt{.}             & \texttt{punct} & 9   & \texttt{beta}  & \_                                         & \texttt{)S} \\
\end{tabular}
}
\caption{Alternative TAGbank format with explicit multiword expression (MWE) annotation and potential support for lexicalized elementary trees} \label{alt-tagbank-format}
\end{subfigure}

\caption{Comparison of the canonical and MWE-annotated TAGbank formats} \label{proposed-tagbank-format}
\end{figure*}

\section{TAG Parsing}

Parsing TAG structures presents unique challenges due to TAG’s extended domain of locality and adjunction mechanism. While TAG is theoretically well-suited for capturing complex syntactic and semantic relations, its parsing complexity has historically limited large-scale empirical research \citep{sarkar-2000-practical}. One major obstacle has been the lack of derivation-based corpora—resources that provide explicit TAG derivation trees—which are essential for training and evaluating both symbolic and statistical TAG parsers.

To address this gap, we introduce TAGbank, a standardized corpus of automatically derived TAG structures aligned with existing syntactic treebanks. Unlike previous resources that offer only derived or surface-level trees, TAGbank provides full derivation histories, enabling the supervision required for learning derivation-based parsers. This corpus makes it possible, for the first time, to compare parsing strategies on a common dataset and to train neural TAG models at scale. It supports systematic benchmarking, cross-formalism comparison, and integration into downstream tasks requiring fine-grained syntactic derivations.

Early research established the theoretical underpinnings of TAG parsing. An Earley-style chart parser for TAG that extended classical parsing techniques to handle adjunction and auxiliary tree composition was proposed \citep{schabes-joshi:1988:ACL}. General parsing strategies for lexicalized grammars, with an emphasis on integrating lexical anchors during syntactic derivation \citep{schabes-abeille-joshi:1988:COLING}.

Efficiency improvements were achieved with the Head-Corner Parsing approach, which predicts head elements prior to full constituent construction \citep{noord:1994:CL}. The Valid Prefix Property provided a theoretical framework for incremental, left-to-right TAG parsing \citep{schabes:1991}. Applied approaches such as the LR-style TAG parser showed that richly lexicalized grammars could be parsed efficiently within a shift-reduce architecture \citep{prolo:2002:EMNLP}. More recent work leveraged generalized chart constraints informed by neural models to prune the search space in TAG and PCFG parsing, improving runtime with no loss in coverage \citep{grnewald-henning-koller:2018:Short}.

Recent work has shown that neural models can effectively handle the structural complexity of TAG. A shift-reduce parser using only 1-best supertag sequences, without relying on lexical information, achieves state-of-the-art performance in unlabeled and labeled attachment scores \citep{kasai-etal-2017-tag}. By embedding supertags as dense vectors, the model captures linguistically meaningful structure and supports the notion that supertagging is almost parsing. Building on this, a graph-based end-to-end parser jointly performs POS tagging, supertagging, and parsing using BiLSTMs, highway connections, and character-level CNNs, surpassing prior benchmarks and demonstrating strong results on tasks such as PETE and Unbounded Dependency Recovery \citep{kasai-etal-2018-end}. These advances confirm TAG’s viability as a formalism for structurally rich language understanding.

In sum, the convergence of symbolic formalism, corpus availability, and neural modeling has positioned TAG parsing as a compelling and theoretically grounded alternative to other lexicalized grammar frameworks. With resources such as TAGbank and recent advances in supertagging and joint parsing architectures, TAG is increasingly well-positioned for integration into modern NLP pipelines that demand both high accuracy and linguistic expressivity.

\section{Conclusion and Future Work}

This paper has presented the design and construction of TAGbank, a corpus of Tree-Adjoining Grammar (TAG) derivations automatically extracted from existing syntactic treebanks. Building on the methodology of CCGbank, our approach enables the transformation of phrase-structure annotations into TAG derivations, making explicit the compositional structure encoded in lexicalized elementary trees. In doing so, TAGbank serves both as a practical resource for computational parsing and as a platform for investigating the linguistic and formal properties of TAG.

We demonstrated the feasibility of large-scale treebank-to-TAG conversion and outlined key challenges in aligning phrase-structure annotations with TAG’s formalism, including lexical anchoring, structural decomposition, and preservation of syntactic function labels. The resulting corpus captures derivational transparency while remaining compatible with surface syntactic annotations.

We plan to extend TAGbank to typologically diverse languages by incorporating the Penn Chinese Treebank \citep{xue-EtAl:2005} and the Penn Korean Treebank \citep{han-EtAl:2002}. 
These languages present distinct structural challenges—such as head-final ordering in Korean and analytic syntax in Chinese—that provide opportunities to evaluate the cross-linguistic robustness of TAG derivations.
A key objective of this extension is to assess whether elementary tree induction and derivation composition can be applied consistently across different annotation schemes and linguistic typologies.

Beyond multilingual grammar extraction, TAGbank also opens up several avenues for broader research. Future work includes integrating semantic annotation layers such as PropBank \citep{palmer-gildea-kingsbury:2005:CL,bonial-EtAl:2015} and AMR \citep{banarescu-EtAl:2013:LAW} on top of TAG derivations, facilitating joint syntactic-semantic analysis within a unified lexicalized framework. Additionally, the corpus may serve as training data for structure-aware large language models that benefit from explicit hierarchical representations, enabling better alignment between symbolic grammar and neural architectures.

Ultimately, TAGbank contributes to the long-term goal of bridging theoretical insights from formal grammar with empirical corpus-based modeling. As TAG continues to be a rich formalism for representing unbounded dependencies and predicate-argument structure \citep{kroch-joshi-1985-linguistic,kroch-1987-unbounded}, and as its generative power remains well-understood in formal language theory \citep{vijayshankar-joshi:1986:HLT,joshi-vijayshanker-weir-1991-convergence,vijayshanker-weir-1994-equivalence}, grounding TAG in multilingual, derivation-based corpora strengthens its role in both linguistic theory and computational practice.


\end{document}